\documentclass[10pt,twocolumn,letterpaper]{article}

\usepackage[pagenumbers]{cvpr} % To force page numbers, e.g. for an arXiv version

\usepackage{multirow}
\usepackage{booktabs, multirow, makecell, xcolor, graphicx}
\usepackage{stfloats}
\usepackage{tabularx,booktabs}
\usepackage{pifont}
\newcommand{\cmark}{\ding{51}}
\newcommand{\xmark}{\ding{55}}

\usepackage{xcolor}

\usepackage{xcolor,booktabs,multirow,makecell,graphicx,amssymb,pifont}
\definecolor{proj}{RGB}{0,0,0}
\newcommand{\p}[1]{#1}  % 推算值，待替换
\newcommand{\projtab}{}         % 整张表为推算值

\definecolor{cvprblue}{rgb}{0.21,0.49,0.74}
\usepackage[pagebackref,breaklinks,colorlinks,allcolors=cvprblue]{hyperref}

\def\confName{CVPR}
\def\confYear{2027}

\title{Wait, Which Look Should Multimodal Reasoning Models Believe?}

\author{Li Ma \quad Yan Zhuang \quad Jiawen Deng\thanks{Corresponding authors.} \quad Fuji Ren\footnotemark[1]\\
University of Electronic Science and Technology of China\\
No. 2006, Xiyuan Ave, West Hi-Tech Zone, Chengdu, Sichuan, 611731, P.R. China\\
{\tt\small \{202421080935, 202211081370\}@std.uestc.edu.cn, \{dengjw, renfuji\}@uestc.edu.cn}
}

\begin{document}
\maketitle
\begin{abstract}
Multimodal reasoning models generate a reasoning trace before the final answer, and frequently look at the image again within this trace. However, later looks are no more accurate than earlier ones. The visual clues from different looks can even contradict each other. For the same image and question, the model states one clue after an early look and a conflicting clue after a later look. When the clues conflict, which look does the model believe? The final answer follows the most recent clue, even when an earlier clue is correct. Telling the model that its clues conflict, without saying which one is correct, reduces the effect of a wrong later clue and improves accuracy substantially. This finding motivates PINC. After the model completes its reasoning, PINC holds back the final answer and inspects the reasoning trace for conflicting clues. If it finds a conflict, PINC notifies the model, which continues the reasoning to a new answer. PINC detects conflicts from the text of the trace alone, so its detector is trained entirely on synthetic text traces. PINC improves accuracy on samples with conflicting clues and on the remaining samples, without modifying the original model.
\end{abstract}    
\section{Introduction}
\label{sec:intro}

\begin{figure}[t]
  \centering
  \includegraphics[width=\linewidth]{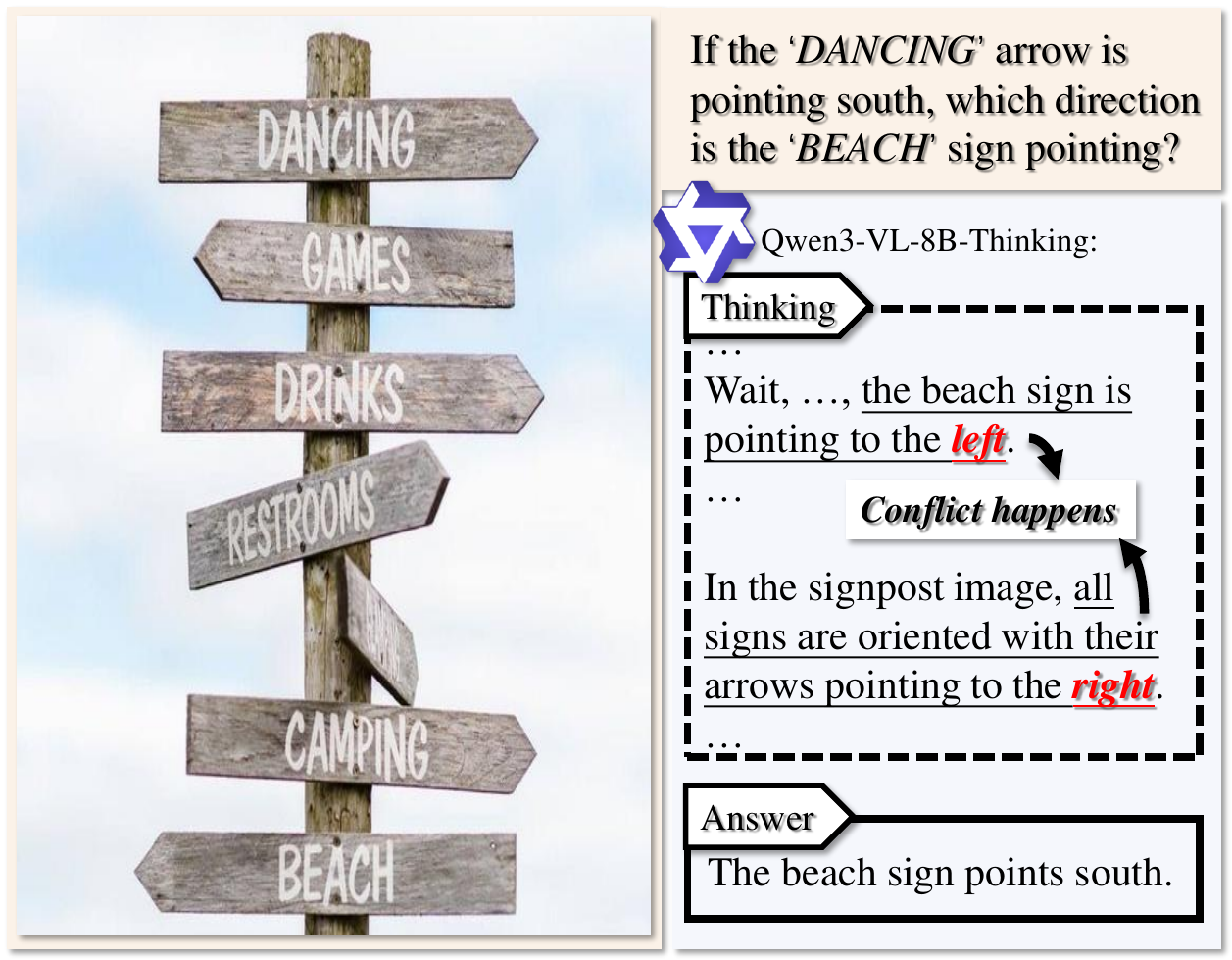}
  \caption{Conflicting visual clues within one reasoning trace. Qwen3-VL-8B-Thinking first states that the beach sign points to the left, which is correct. After a later look, it states that all signs point to the right, which conflicts with the earlier clue. The final answer follows the later clue and is wrong.}
  \label{fig:teaser}
\end{figure}

Multimodal reasoning models generate a reasoning trace before giving the final answer~\cite{qwen3vl,glm41v,kimivl}, and they frequently look at the image again within this trace~\cite{visualswap}. Looking again, however, brings little new visual information, because the model tends to reuse the reasoning it has already written~\cite{recompute}. Our analysis shows that the visual clues from different looks can even contradict each other. As shown in Fig.~\ref{fig:teaser}, the model first states that the beach sign points to the left, but after a later look, it states that all signs point to the right. Across four multimodal reasoning models~\cite{qwen3vl,kimivl,glm41v} and four benchmarks~\cite{cvbench,blink,naturalbench,tallyqa}, 42\% of the samples contain such conflicting clues, and later looks are no more accurate than earlier ones. This raises a question: when the visual clues conflict, which look does the model believe?

To answer this question, we first examine the original traces. When a correct clue is followed by a conflicting wrong clue, the final answer agrees with the wrong clue in 68.9\% of the samples, almost three times as often as with the correct clue (24.6\%). To test whether the order of the clues itself affects the answer, we take from each trace a pair of conflicting clues that the model wrote, one correct and one wrong, append the two clues to the end of the trace in both orders, and let the model continue to the final answer. The answer follows the clue that comes last: accuracy is 82.8\% when the correct clue comes last and 20.9\% when the wrong clue comes last. A clue in the earlier position can still determine the answer: when we replace the later clue with a clue that is unrelated to the question, the answer follows the earlier clue.

We further find that pointing out the conflict reduces the effect of a wrong later clue. We append one sentence after the two clues, which tells the model that its clues disagree but does not say which clue is correct. When the wrong clue comes last, accuracy rises from 20.9\% to 66.6\%. When the correct clue comes last, accuracy remains at 81.8\% on average, so the sentence does not simply make the model switch to the other clue.

Based on this finding, we propose PINC, whose four letters stand for its four steps: Pause, Inspect, Notify, and Continue. PINC pauses before the final answer is returned and inspects the reasoning trace with a detector. If the detector finds conflicting clues, PINC notifies the model with a sentence that states the object and the attribute on which the clues disagree, and the unmodified model continues to a new answer. Otherwise, the original answer is returned.

The conflicting clues are written in the trace, so the detector reads only text and can be trained entirely on synthetic traces. Claude Opus 5.5~\cite{claudeopus} imagines 1,000 scenes, writes correct and wrong visual clues for each scene, and then writes reasoning traces from these clues. The clues determine the labels, so no manual annotation is needed. The detector consists of two small models: one locates the sentences that conflict, and the other extracts the object and the attribute on which they conflict. Across four models and four benchmarks, PINC raises accuracy by 8.4 points on the samples with conflicting clues. It also improves accuracy on the remaining samples, because a new trace for the same question can still contain a conflict.

Our contributions are summarized as follows:
\begin{itemize}
    \item We find that the visual clues from different looks conflict in 42\% of the samples, and that later looks are no more accurate than earlier ones.
    \item We show that a later clue overrides an earlier one, even when the earlier clue is correct, and that a hint pointing out the conflict reduces this effect.
    \item We propose PINC, which detects conflicting clues from the text of the trace and notifies the model. Its detector is trained only on synthetic traces, and the original model is not modified.
\end{itemize}

\begin{figure*}[t]
  \centering
  \begin{subfigure}[t]{0.45\linewidth}
    \centering
    \includegraphics[width=\linewidth]{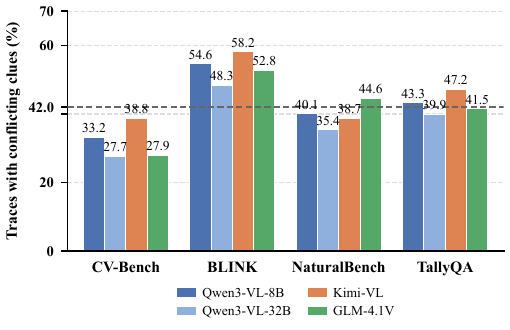}
    \caption{}
  \end{subfigure}
  \hfill
  \begin{subfigure}[t]{0.45\linewidth}
    \centering
    \includegraphics[width=\linewidth]{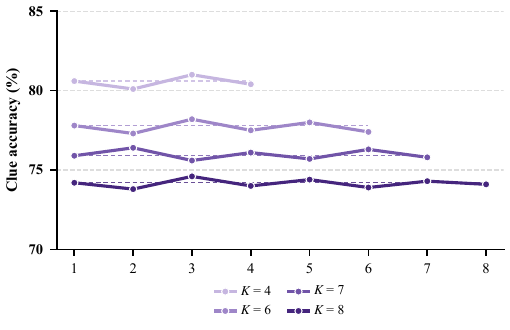}
    \caption{}
  \end{subfigure}
  \caption{(a)~Share of samples with at least one pair of conflicting clues. The dashed line marks the overall share. (b)~Accuracy of the clue from the $k$-th look, grouped by the number of looks $K$. Dashed lines mark the first look.}
  \label{fig:conflict}
\end{figure*}
\section{Related Work}
\label{sec:related}

\paragraph{Looking again during reasoning.}
Multimodal reasoning models attend less to the image as their reasoning grows longer~\cite{morethinking,lookagain,vre}. VisualSwap~\cite{visualswap} replaces the image after the model has reasoned over it and finds that the model rarely notices the change. Recompute or Reuse~\cite{recompute} shows that the model reuses the reasoning it has already written instead of reading the current image. Other studies make looking again more useful through decoding strategies or additional visual computation~\cite{lookback_uncertainty,visref,asktwice,blinktwice}. These studies examine whether a later look uses the image. We examine whether the visual clues from different looks agree, and which clue the answer follows when they do not.

\paragraph{Conflicts in reasoning.}
Prior work studies conflicts between the reasoning of a model and information that the experimenter introduces: a swapped image~\cite{visualswap,visualflip}, a perturbation injected into the reasoning~\cite{logicgraph}, a different knowledge source~\cite{knowledgeconflict}, updated evidence~\cite{traceevidence}, or misleading text in a video~\cite{visualtrust}. In text-only reasoning, language models also favor the steps closest to the conclusion~\cite{proofteller}. The conflicts we study arise within a single trace on an unchanged image, between clues that the model itself wrote, and we separate the order of the clues from their content by appending the same pair in both orders.

\paragraph{Self-correction for multimodal reasoning.}
Many methods train the reasoning model to reflect on and correct its own reasoning, with reinforcement learning~\cite{vlrethinker,lookback,lookagain,vrrl,reflectr1,cofire} or supervised fine-tuning~\cite{mirror}, and ReVISE~\cite{revise} trains a model to detect inconsistencies between tool outputs and the image. Training-free methods instead reuse earlier reasoning across samples~\cite{hgivr} or intervene in decoding~\cite{lookback_uncertainty,recompute}. All these methods change how the reasoning model is trained or how it decodes. PINC leaves the reasoning model unchanged and trains only a small text-only detector on synthetic traces.
\section{Visual Clue Conflicts in Reasoning Traces}
\label{sec:analysis}

A multimodal reasoning model often looks at the image several times within one reasoning trace, and each time it writes down visual clues, statements about what it sees. We study what happens when these clues conflict: how often clues from different looks conflict (\S\ref{sec:prevalence}), and which clue the final answer follows when they do (\S\ref{sec:selection}).

\subsection{Analysis Setup}
\label{sec:setup}

\paragraph{Models.}
The analysis covers four multimodal reasoning models: Qwen3-VL-8B-Thinking (Qwen3-VL-8B), Qwen3-VL-32B-Thinking (Qwen3-VL-32B)~\cite{qwen3vl}, Kimi-VL-A3B-Thinking(Kimi-VL)~\cite{kimivl}, and GLM-4.1V-9B-Thinking (GLM-4.1V)~\cite{glm41v}. The names in parentheses are used in the rest of the paper. The four models come from three model families, and all of them output the full reasoning trace.

\paragraph{Benchmarks.}
Questions come from four benchmarks: CV-Bench~\cite{cvbench}, the validation split of BLINK~\cite{blink}, the English set of NaturalBench~\cite{naturalbench}, and the complex test split of TallyQA~\cite{tallyqa}. These benchmarks require close inspection of the image. They cover spatial relations, counting, and fine-grained attributes, and they include both multiple-choice questions and open-ended counting. Appendix~A lists the number of questions per benchmark.

\paragraph{Conflicts.} The conflicts we study are pairs of visual clues that describe the same attribute of the same object in incompatible ways, while the trace keeps both clues as what the model sees. In Fig.~\ref{fig:teaser}, the model first states that the beach sign points to the left and later states that it points to the right. We distinguish such conflicts from two other cases in which a trace contains incompatible statements, and we do not count either case as a conflict. The first case is an explicit revision, in which a later statement marks an earlier clue as wrong, as in ``Actually, the sign points to the left, not to the right.'' Explicit revisions appear in 9\% of the samples. The second case is a hypothetical statement, in which the model describes a possibility rather than what it sees, as in ``If the sign pointed to the right, the beach would be behind us.'' Hypothetical statements appear in 21\% of the samples.

\subsection{Prevalence of Conflicting Clues}
\label{sec:prevalence}

We first examine what the model writes each time it looks at the image. Each reasoning trace is split into sentences, and each sentence forms one segment. Claude Opus 5.5 then reads the segments, extracts the visual clues, and judges whether any two clues conflict under the definition in \S\ref{sec:setup}. This judgment uses only the text of the trace. To verify it, human annotators check a random subset of 800 traces, stratified by model and benchmark, and they agree with Claude on 98.7\% of them. Human annotators also label every clue as correct or incorrect by comparing it with the image. Appendix~B describes the full procedure.

Looking again rarely brings new information. Almost every sample (91\%) contains at least one later look that repeats an earlier clue, while only 20\% of the samples contain a look that adds information. Later looks do not only repeat earlier clues. They can also contradict them.

\paragraph{Finding 1: Conflicting clues appear in 42\% of the samples.}
In 42\% of the samples, the trace contains at least one pair of conflicting clues (Fig.~\ref{fig:conflict}a). The share ranges from 37.8\% to 45.7\% across the four models and from 31.9\% to 53.5\% across the four benchmarks. Conflicts therefore appear on every model and every benchmark.

\paragraph{Finding 2: Later looks are no more accurate than earlier ones.}
If later looks corrected earlier mistakes, the clues would become more accurate as the model looks more often. To test this, we take from each look the clue about the attribute that the question asks for, and we measure how often this clue is correct. Traces with more looks tend to come from harder questions, so traces with different numbers of looks cannot be compared directly. We therefore group the traces by their number of looks $K$ and compare the looks within each group. Fig.~\ref{fig:conflict}b shows the four most frequent groups, $K = 4, 6, 7, 8$. Within each group, clue accuracy stays within 1 point from the first look to the last, and the last look is not above the first one (dashed lines). This stable accuracy does not mean that the clues stay the same. Among all pairs of consecutive looks, the clue changes from correct to wrong in 4.2\% and from wrong to correct in 4.0\%. Later looks correct some mistakes but introduce about as many new ones.

In the four most frequent groups, later clues are thus no more accurate than earlier ones, yet they often conflict with them. Which clue the model follows then decides whether its answer is correct. \S\ref{sec:selection} examines this choice.

\subsection{Answer Selection under Conflicting Clues}
\label{sec:selection}

To analyze how the model chooses between conflicting clues, we select the samples whose traces contain conflicting clues (\S\ref{sec:prevalence}). In 74\% of these samples, one of the conflicting clues supports the ground-truth answer and another supports a wrong answer. We call them clue A and clue B. In another 7.2\%, both clues are wrong, and in the remaining 18.8\%, the conflict does not concern the attribute that the question asks for. We first examine which clue the final answer agrees with in the original traces (Table~\ref{tab:natural}). The answer agrees with the later clue in 68.9\% of the samples in which the wrong clue comes later, and in 81.4\% of the samples in which the correct clue comes later. When both clues are wrong, the answer also agrees with the later clue more often than with the earlier one, in 60.2\% against 18.6\% of the samples. This agreement alone does not show that the order of the clues affects the answer, because the later clue may only reflect an answer that the model already prefers.

To test the effect of the order itself, we run controlled experiments on the samples with clue A and clue B. From each trace, we also take a neutral clue, a correct clue about another part of the image that does not bear on the question. The neutral clue is chosen to have a similar length to the clue it replaces. All three clues are sentences that the model itself wrote in this trace. None of them is written by hand or by another model.

Each experiment keeps the full reasoning trace, appends two of these clues at its end, and lets the model continue to the final answer. Some experiments also append a conflict hint after the clues:
\begin{quote}
\textit{Wait, my clues of [object and attribute] do not agree. Let me check again.}
\end{quote}
The bracket is filled with the object and the attribute that Claude Opus 5.5 identifies when it judges the conflict (\S\ref{sec:prevalence}). The hint states where the clues conflict, but it does not say which clue is correct. Table~\ref{tab:prefill} reports all results, and the numbers below are averages over the four models and the four benchmarks.

\begin{table}[t]
\centering
\caption{Final answers on the original traces with conflicting clues (\%). \cmark: correct clue. \xmark: wrong clue. Arrows give the order of the two clues in the trace.}
\label{tab:natural}
\small
\setlength{\tabcolsep}{2pt}
\begin{tabular*}{\linewidth}{@{\extracolsep{\fill}}ccccc@{}}
\toprule
 & & \multicolumn{3}{c}{Final answer agrees with} \\
\cmidrule(lr){3-5}
Clue order & Proportion & Earlier clue & Later clue & Neither \\
\midrule
\cmark\,$\rightarrow$\,\xmark & 39.0 & 24.6 & \textbf{68.9} & 6.5 \\
\xmark\,$\rightarrow$\,\cmark & 35.0 & 13.8 & \textbf{81.4} & 4.8 \\
\xmark\,$\rightarrow$\,\xmark & 7.2 & 18.6 & \textbf{60.2} & 21.2 \\
\bottomrule
\end{tabular*}
\end{table}

\begin{table}[t]
  \centering
  \caption{Accuracy (\%) with two appended clues. A: correct clue. B: wrong clue. Neutral: correct but unrelated clue.}
  \label{tab:prefill}
  \small
  \setlength{\tabcolsep}{3pt}
  \begin{tabular}{l c ccccc}
    \toprule
    Appended & Hint & \makecell{Qwen3-\\VL-8B} & \makecell{Qwen3-\\VL-32B} & \makecell{Kimi-\\VL} & \makecell{GLM-\\4.1V} & Avg. \\
    \midrule
    A + B       & & 20.6 & 23.9 & 17.8 & 21.3 & 20.9 \\
    B + A       & & 82.7 & 85.4 & 79.6 & 83.5 & 82.8 \\
    \midrule
    A + neutral & & 89.4 & 91.2 & 86.7 & 90.1 & 89.4 \\
    B + neutral & & 13.8 & 12.1 & 16.5 & 13.2 & 13.9 \\
    \midrule
    A + B       & \checkmark & 66.3 & 70.8 & 61.9 & 67.4 & 66.6 \\
    B + A       & \checkmark & 81.5 & 84.6 & 78.2 & 82.9 & 81.8 \\
    \bottomrule
  \end{tabular}
\end{table}

\begin{table}[t]
  \centering
  \caption{Accuracy (\%) on the original traces with conflicting clues, before $\rightarrow$ after the conflict hint.}
  \label{tab:hint}
  \footnotesize
  \setlength{\tabcolsep}{4.5pt}
  \begin{tabular}{l cccc}
    \toprule
    Benchmark & \makecell{Qwen3-\\VL-8B} & \makecell{Qwen3-\\VL-32B} & \makecell{Kimi-\\VL} & \makecell{GLM-\\4.1V} \\
    \midrule
    CV-Bench     & 65.1$\rightarrow$78.9 & 68.6$\rightarrow$80.2 & 62.4$\rightarrow$77.4 & 66.7$\rightarrow$79.0 \\
    BLINK        & 54.2$\rightarrow$63.5 & 60.4$\rightarrow$68.7 & 41.0$\rightarrow$51.3 & 56.5$\rightarrow$65.1 \\
    NaturalBench & 25.6$\rightarrow$30.5 & 28.8$\rightarrow$33.2 & 26.8$\rightarrow$32.4 & 21.2$\rightarrow$25.9 \\
    TallyQA      & 60.7$\rightarrow$71.9 & 61.9$\rightarrow$71.6 & 59.7$\rightarrow$71.6 & 59.9$\rightarrow$70.3 \\
    \midrule
    Avg.         & 51.4$\rightarrow$61.2 & 54.9$\rightarrow$63.4 & 47.5$\rightarrow$58.2 & 51.1$\rightarrow$60.1 \\
    \bottomrule
  \end{tabular}
\end{table}

\paragraph{Finding 3: A later clue overrides an earlier one, even when the earlier clue is correct.}
We first test whether the order of the clues decides the answer. We append A and B in both orders, so that the two conditions contain the same clues and differ only in which clue comes last. Accuracy is 20.9\% when the wrong clue B comes last (A\,+\,B) and 82.8\% when the correct clue A comes last (B\,+\,A). In both orders, most answers follow the last clue. The answer therefore depends on the order of the two clues. The result is similar when the two appended clues are reworded instead of copied (21.7\% and 82.6\%, Appendix~E), so it does not depend on repeating the original sentences.

One possible reason for this behavior is that the model does not see the earlier clue when it writes its conclusion. If this were the case, a clue in the first position will not affect the answer. To test this, we replace the second clue with the neutral clue, which is unrelated to the question and has a similar length. The first clue therefore stays at the same distance from the conclusion. The answer then follows the first clue: accuracy is 89.4\% when the first clue is A and 13.9\% when it is B. A clue in the first position can therefore determine the answer. The earlier clue loses its effect only in A\,+\,B and B\,+\,A, where a conflicting clue follows it. Since the distance is the same in all four conditions, the conflict is what removes the effect of the earlier clue. The later clue overrides the earlier one, even when the earlier clue is correct.

\paragraph{Finding 4: A conflict hint reduces the effect of a wrong later clue.}
Finding 3 raises the question of whether the effect of a wrong later clue can be reduced without telling the model which clue is correct. To test this, we append the conflict hint after the clues. The hint states where the clues conflict but not which clue is correct, so any gain does not come from the hint revealing the answer.

With the hint, accuracy for A\,+\,B rises from 20.9\% to 66.6\%, while accuracy for B\,+\,A changes little, from 82.8\% to 81.8\%. The hint does not simply make the model switch to the other clue. If it did, accuracy for B\,+\,A would fall sharply. Instead, the answer moves to the correct clue when the wrong clue comes last, and it stays with the correct clue when the correct clue comes last.

The same hint also helps on the original traces. It raises accuracy by 9.5 points on average across the four models and four benchmarks (Table~\ref{tab:hint}).

Together, the two findings show that the answer depends on the order of conflicting clues, and that a hint pointing out the conflict reduces this dependence. In this analysis, the hint is built from annotated conflicts. PINC detects the conflicts automatically and appends the same hint (\S\ref{sec:method}).

% TODO: to be written.

\begin{figure*}[t]
  \centering
  \includegraphics[width=\linewidth]{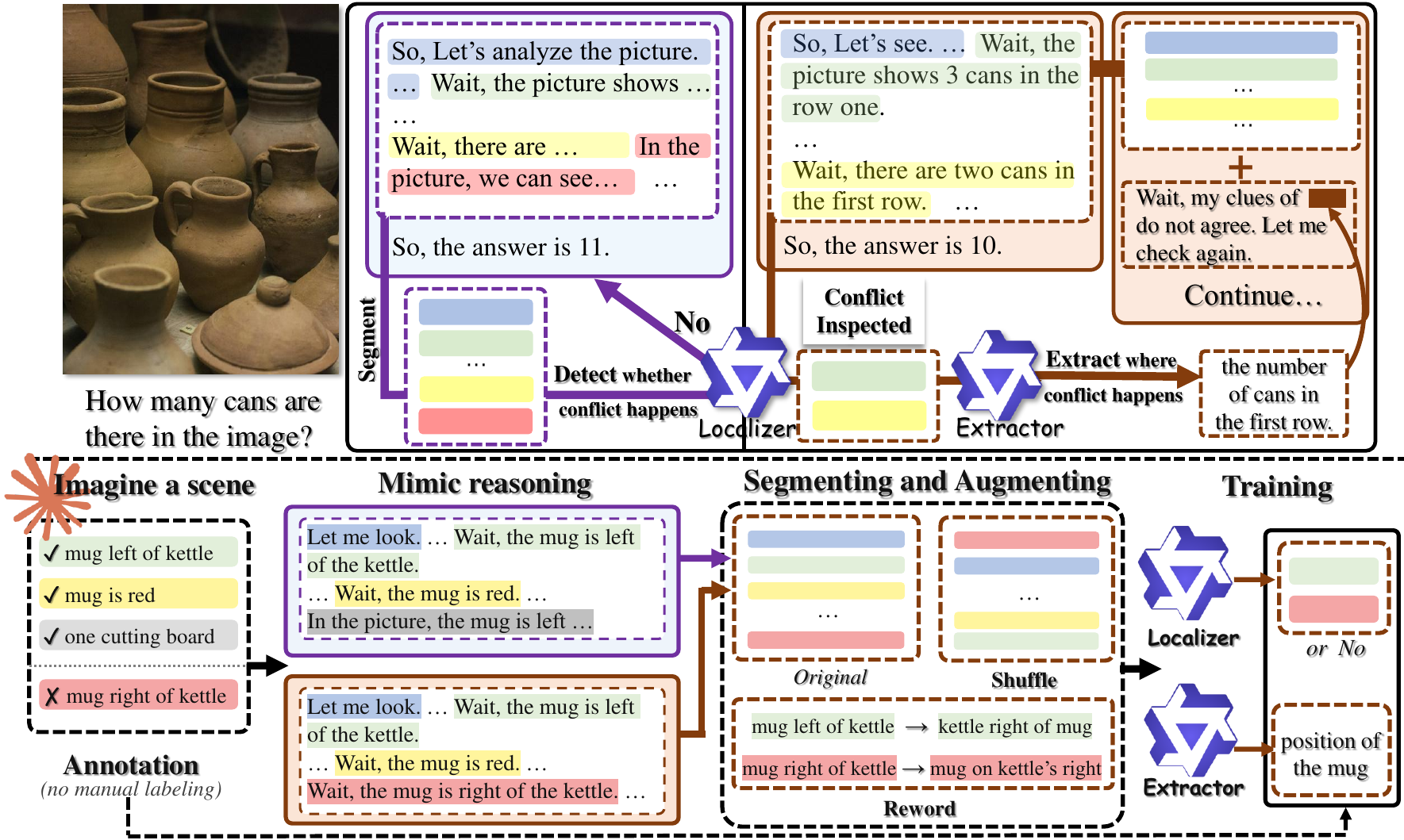}
  \caption{Overview of PINC. Top: inference. The localizer checks the segments of the trace, the extractor extracts the conflicting object and attribute, and the model continues from the hint. Bottom: training of the two models on synthetic traces written by Claude Opus 5.5.}
  \label{fig:method}
\end{figure*}

\section{PINC}
\label{sec:method}

The analysis in \S\ref{sec:analysis} leads to two design choices. First, a hint that points out the conflict reduces the effect of a wrong later clue (Finding 4). The method therefore only needs to point out the conflict and can leave the answer to the model. Second, the conflicting clues are written in the trace, so a detector can find them from the text alone. PINC follows both choices in four steps: Pause, Inspect, Notify, and Continue (Fig.~\ref{fig:method}).

\subsection{Inference Pipeline}
\label{sec:pipeline}

\paragraph{Pause.}
The model first generates its complete output, which consists of the reasoning trace and the final answer. PINC takes the reasoning trace, which ends right before the final answer, and holds the answer back until the trace has been inspected.

\paragraph{Inspect.}
The trace is split into sentences, and each sentence forms one segment. The detector, which consists of two text-only models, then reads the segments. The localizer reads all segments and outputs the indices of the segments whose clues conflict, or ``no'' if there is none. The extractor reads the conflicting segments and extracts the object and the attribute on which they disagree, for example ``the number of cans in the first row''.

\paragraph{Notify.}
If the localizer outputs ``no'', PINC returns the original answer. Otherwise, PINC discards this answer and appends the conflict hint from \S\ref{sec:selection} to the end of the trace, with the bracket filled by the output of the extractor. When the trace contains several conflicts, the hint lists all of them. As in \S\ref{sec:selection}, the hint does not say which clue is correct.

\paragraph{Continue.}
The model continues its reasoning from the hint and generates a new final answer. Its parameters are not changed. The trace before the hint is kept in the KV cache, so only the hint and the new text are computed.

\begin{table*}[t]
\centering
\caption{Accuracy (\%) on the conflict subset (Conf.), the other
subset (Other), and all samples (Ovr.). Values in parentheses
in the PINC rows indicate absolute improvements in percentage
points over Original for the same model.
Generic hint: a hint without a detector.
SC: self-consistency with three samples.}
\label{tab:main}
\resizebox{\linewidth}{!}{%
\begin{tabular}{l l lll lll lll lll}
\toprule
& & \multicolumn{3}{c}{CV-Bench} & \multicolumn{3}{c}{BLINK} & \multicolumn{3}{c}{NaturalBench} & \multicolumn{3}{c}{TallyQA} \\
\cmidrule(lr){3-5} \cmidrule(lr){6-8} \cmidrule(lr){9-11} \cmidrule(lr){12-14}
Model & Method & Conf. & Other & Ovr. & Conf. & Other & Ovr. & Conf. & Other & Ovr. & Conf. & Other & Ovr. \\
\midrule
\multirow{4}{*}{Qwen3-VL-8B}
& Original & 65.8 & 90.5 & 82.3 & 55.0 & 70.0 & 61.8 & 26.0 & 40.0 & 34.4 & 61.6 & 83.0 & 73.7 \\
& Generic hint & 70.6 & 89.3 & 83.1 & 57.6 & 67.6 & 62.1 & 27.4 & 38.3 & 33.9 & 65.7 & 81.4 & 74.6 \\
& SC & 68.4 & 91.1 & 83.6 & 58.0 & 71.3 & 64.0 & 27.8 & 40.9 & 35.7 & 64.7 & 83.9 & 75.6 \\
& PINC & 78.1\,\textcolor{gray}{\scriptsize (+12.3)} & 91.7\,\textcolor{gray}{\scriptsize (+1.2)} & 87.2\,\textcolor{gray}{\scriptsize (+4.9)} & 62.9\,\textcolor{gray}{\scriptsize (+7.9)} & 71.6\,\textcolor{gray}{\scriptsize (+1.6)} & 66.9\,\textcolor{gray}{\scriptsize (+5.1)} & 30.4\,\textcolor{gray}{\scriptsize (+4.4)} & 40.8\,\textcolor{gray}{\scriptsize (+0.8)} & 36.6\,\textcolor{gray}{\scriptsize (+2.2)} & 71.3\,\textcolor{gray}{\scriptsize (+9.7)} & 84.3\,\textcolor{gray}{\scriptsize (+1.3)} & 78.7\,\textcolor{gray}{\scriptsize (+5.0)} \\
\midrule
\multirow{4}{*}{Qwen3-VL-32B}
& Original & 69.1 & 92.7 & 86.2 & 61.5 & 76.0 & 69.0 & 29.1 & 44.0 & 38.7 & 62.5 & 83.5 & 75.1 \\
& Generic hint & 73.7 & 91.9 & 86.9 & 64.6 & 74.5 & 69.7 & 30.8 & 42.7 & 38.5 & 66.7 & 82.5 & 76.2 \\
& SC & 71.2 & 93.1 & 87.1 & 64.7 & 76.9 & 71.0 & 31.1 & 44.7 & 39.9 & 65.1 & 84.1 & 76.5 \\
& PINC & 79.5\,\textcolor{gray}{\scriptsize (+10.4)} & 93.5\,\textcolor{gray}{\scriptsize (+0.8)} & 89.6\,\textcolor{gray}{\scriptsize (+3.4)} & 68.5\,\textcolor{gray}{\scriptsize (+7.0)} & 77.2\,\textcolor{gray}{\scriptsize (+1.2)} & 73.0\,\textcolor{gray}{\scriptsize (+4.0)} & 33.1\,\textcolor{gray}{\scriptsize (+4.0)} & 44.6\,\textcolor{gray}{\scriptsize (+0.6)} & 40.5\,\textcolor{gray}{\scriptsize (+1.8)} & 71.3\,\textcolor{gray}{\scriptsize (+8.8)} & 84.5\,\textcolor{gray}{\scriptsize (+1.0)} & 79.2\,\textcolor{gray}{\scriptsize (+4.1)} \\
\midrule
\multirow{4}{*}{Kimi-VL}
& Original & 63.5 & 88.0 & 78.5 & 42.3 & 56.0 & 48.0 & 27.4 & 42.0 & 36.4 & 60.9 & 81.0 & 71.5 \\
& Generic hint & 67.4 & 85.9 & 78.7 & 44.7 & 53.0 & 48.1 & 28.6 & 40.1 & 35.7 & 64.4 & 78.6 & 71.9 \\
& SC & 66.7 & 88.9 & 80.3 & 44.2 & 57.7 & 49.8 & 29.0 & 43.1 & 37.7 & 64.3 & 82.3 & 73.8 \\
& PINC & 76.7\,\textcolor{gray}{\scriptsize (+13.2)} & 89.5\,\textcolor{gray}{\scriptsize (+1.5)} & 84.5\,\textcolor{gray}{\scriptsize (+6.0)} & 51.1\,\textcolor{gray}{\scriptsize (+8.8)} & 57.9\,\textcolor{gray}{\scriptsize (+1.9)} & 53.9\,\textcolor{gray}{\scriptsize (+5.9)} & 32.2\,\textcolor{gray}{\scriptsize (+4.8)} & 42.9\,\textcolor{gray}{\scriptsize (+0.9)} & 38.8\,\textcolor{gray}{\scriptsize (+2.4)} & 71.5\,\textcolor{gray}{\scriptsize (+10.6)} & 82.6\,\textcolor{gray}{\scriptsize (+1.6)} & 77.4\,\textcolor{gray}{\scriptsize (+5.9)} \\
\midrule
\multirow{4}{*}{GLM-4.1V}
& Original & 67.3 & 92.0 & 85.1 & 57.5 & 72.0 & 64.3 & 21.7 & 33.0 & 28.0 & 60.6 & 82.0 & 73.1 \\
& Generic hint & 71.0 & 90.7 & 85.2 & 60.4 & 69.9 & 64.8 & 23.2 & 31.6 & 27.9 & 64.2 & 80.1 & 73.5 \\
& SC & 69.6 & 92.5 & 86.1 & 60.6 & 73.2 & 66.5 & 22.9 & 33.6 & 28.9 & 63.5 & 83.0 & 74.9 \\
& PINC & 77.9\,\textcolor{gray}{\scriptsize (+10.6)} & 92.9\,\textcolor{gray}{\scriptsize (+0.9)} & 88.7\,\textcolor{gray}{\scriptsize (+3.6)} & 65.0\,\textcolor{gray}{\scriptsize (+7.5)} & 73.4\,\textcolor{gray}{\scriptsize (+1.4)} & 69.0\,\textcolor{gray}{\scriptsize (+4.7)} & 26.1\,\textcolor{gray}{\scriptsize (+4.4)} & 33.9\,\textcolor{gray}{\scriptsize (+0.9)} & 30.4\,\textcolor{gray}{\scriptsize (+2.4)} & 69.8\,\textcolor{gray}{\scriptsize (+9.2)} & 83.2\,\textcolor{gray}{\scriptsize (+1.2)} & 77.6\,\textcolor{gray}{\scriptsize (+4.5)} \\
\bottomrule
\end{tabular}}
\end{table*}

\subsection{Training the Detector}
\label{sec:detector}

Both models of the detector are Qwen3-4B~\cite{qwen3} with the thinking mode turned off. Since they read only text, their training data does not need images. We build this data from synthetic traces.

\paragraph{Synthetic traces.}
Claude Opus 5.5 imagines 1000 scenes. For each scene, it writes correct visual clues, together with wrong clues that contradict them. It then writes reasoning traces from these clues in the style of a multimodal reasoning model, including the step-by-step logic and the repetitions that such traces contain. Each trace that uses only correct clues is paired with a trace in which wrong clues are mixed into the correct ones. The two traces in a pair differ only in whether their clues conflict. The traces are split into sentences in the same way as at inference.

\paragraph{Training targets.}
The clues used to write a trace determine its labels, so no manual annotation is needed. The localizer is trained to output the indices of the conflicting segments for a trace with wrong clues, and ``no'' for a trace without them. The extractor is trained to output the object and the attribute on which the conflicting segments disagree. We use two models instead of one because the two tasks differ. Locating a conflict requires comparing all segments of a long trace, while extracting it requires reading only the conflicting segments. \S\ref{sec:detector_analysis} compares this design with a single model.

\paragraph{Removing order and wording bias.}
A detector trained on these traces could learn two shortcuts. It could rely on where the conflicting clue appears, for example near the end of the trace. It could also rely on how the clue is phrased. To remove the order bias, we shuffle the segments of each trace, which does not change whether the trace contains a conflict. To remove the wording bias, we reword the clues, so that a conflicting pair stays conflicting and a consistent pair stays consistent.
\section{Experiments}
\label{sec:experiments}

\subsection{Setup}
\label{sec:exp_setup}

We evaluate PINC on the four models and four benchmarks from \S\ref{sec:setup}. Results are reported on three sets of samples. The conflict subset contains the samples whose traces contain conflicting clues in \S\ref{sec:prevalence}, the other subset contains the remaining samples, and the full set combines the two. For evaluation, each model generates a new reasoning trace and answer for every sample, with temperature 0.6 and top-p 0.95. Original denotes this answer. The generic hint and PINC are applied to the same trace, so the three methods differ only in what is appended to it. Self-consistency uses this trace and two additional samples. The two subsets are fixed by the annotation in \S\ref{sec:prevalence} and are the same for all methods. For NaturalBench, we follow its group accuracy~\cite{naturalbench}. A group consists of two images and two questions, and a sample counts as correct only when all four answers of its group are correct. The two subsets are defined per sample as on the other benchmarks, and the accuracy over all samples equals the standard group accuracy. We repeat the evaluation three times and report the mean. Appendix~C gives the standard deviations. 

We compare PINC with two baselines. The generic hint appends the sentence ``Wait, let me check whether my clues conflict.'' to every trace, without a detector. Self-consistency (SC)~\cite{selfconsistency} samples three traces for each question and takes the majority answer.

\subsection{Main Results}
\label{sec:main_results}

\paragraph{PINC improves accuracy on samples with conflicting clues.}
On the conflict subset, PINC raises accuracy by 8.4 points on average across the four models and four benchmarks (Table~\ref{tab:main}), and the gain appears for every model and benchmark. For reference, the hint with annotated conflicts raises accuracy by 9.5 points on the traces analyzed in \S\ref{sec:selection} (Table~\ref{tab:hint}). PINC reaches a similar gain without any annotation.

\paragraph{PINC also improves accuracy on the other samples.}
On the other subset, accuracy rises by 1.2 points. These samples had no conflict in \S\ref{sec:prevalence}, but a new trace for the same question can still contain one. The detector reports a conflict in 16.7\% of the new traces on this subset, and PINC notifies the model in these cases. In the remaining cases, PINC returns the original answer.

\paragraph{Telling the model where the conflict is matters.}
The generic hint asks the model to look for conflicts by itself. It recovers less than half of the gain of PINC on the conflict subset, and it lowers accuracy on the other subset by 0.8 to 3.0 points, because it is also added to traces that contain no conflict. Self-consistency improves the two subsets only slightly, by 2.5 and 0.9 points on average, at three times the generation cost. Over all samples, the generic hint and self-consistency raise accuracy by 0.3 and 1.6 points, compared with 4.1 points for PINC.

\subsection{Detector Analysis}
\label{sec:detector_analysis}

The detector is evaluated on a test set that combines two sources. One source is new synthetic traces that Claude Opus 5.5 writes in the same way as the training data, with no overlap. The other source is the real traces from \S\ref{sec:prevalence}, both with and without conflicting clues. We report three metrics. Localization F1 measures how well the localizer finds the conflicting segments. Extraction accuracy measures how often the extractor outputs the correct object and attribute when it is given the correct segments. End-to-end accuracy measures how often the full detector outputs the correct object and attribute, or ``no'' for a trace without conflict.

\begin{figure}[t]
  \centering
  \includegraphics[width=0.9\linewidth]{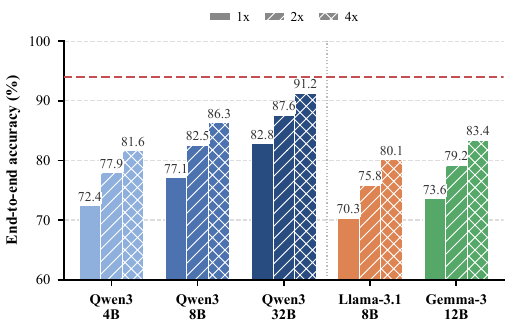}
  \caption{End-to-end accuracy of single-model detectors with 1$\times$, 2$\times$, and 4$\times$ training data. The dashed line marks PINC (two Qwen3-4B models, 1$\times$ data).}
  \label{fig:split}
\end{figure}

\paragraph{Two small models outperform one large model.}
A single model could also read the whole trace and output the conflicting object and attribute directly. To compare the two designs, we train single-model detectors from three families: Qwen3 at 4B, 8B, and 32B, Llama-3.1-8B~\cite{llama3}, and Gemma-3-12B~\cite{gemma3}. Each single model is trained with one, two, and four times the training data of PINC. The end-to-end accuracy of a single model rises with both its size and its data (Fig.~\ref{fig:split}). The best single model, Qwen3-32B with four times the data, reaches 91.2\%. PINC reaches 94.0\% with two Qwen3-4B models and the original data. Llama-3.1-8B and Gemma-3-12B show the same trend at lower accuracy.

We also use Claude Opus 5.5 as a detector without any training. It reaches an end-to-end accuracy of 95.1\%, close to the 94.0\% of PINC. The two Qwen3-4B models thus come close to a much larger model on this task, and they run locally.

\begin{figure}[t]
  \centering
  \includegraphics[width=0.8\linewidth]{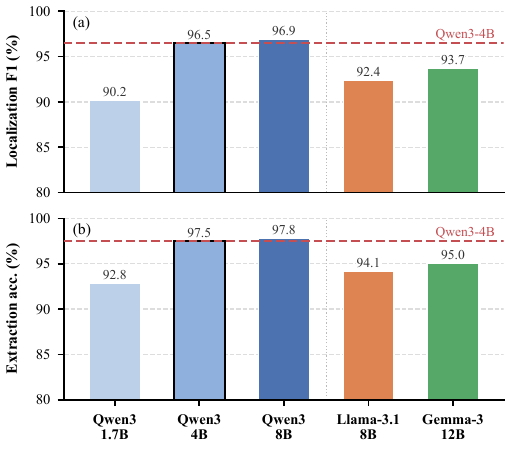}
  \caption{(a)~Localization F1 of the localizer. (b)~Extraction accuracy of the extractor. The dashed line marks Qwen3-4B.}
  \label{fig:size}
\end{figure}

\paragraph{Qwen3-4B is sufficient for both models.}
To choose the model for each step, we replace one model at a time and keep the other as Qwen3-4B. All candidates use the same training data, prompt, and training settings. Within Qwen3, accuracy rises from 1.7B to 4B, and 8B adds less than 0.5 points on both steps (Fig.~\ref{fig:size}). Llama-3.1-8B and Gemma-3-12B are larger than Qwen3-4B, yet they score lower on both steps.

\paragraph{Removing order and wording bias is necessary.}
Table~\ref{tab:detector} reports the detection rate, the share of conflicting traces that the detector flags, and the false-alarm rate, the share of traces without conflict that it flags. Without shuffled segments, the false-alarm rate rises from 4.8\% to 11.7\%. Without reworded clues, the detection rate drops from 95.8\% to 88.4\%. Without both, end-to-end accuracy falls from 94.0\% to 82.7\%.

\paragraph{Both parts of the hint contribute.}
The hint of PINC tells the model two things: that its clues conflict, and on which object and attribute they conflict. The gain of PINC could therefore come from pointing the model to the attribute alone. To separate the two parts, we keep the localizer and change only the hint that is appended when it reports a conflict (Table~\ref{tab:hint_ablation}). The hint with only the conflict reads ``Wait, my clues do not agree. Let me check again.'' The hint with only the object and the attribute reads ``Let me look at [object and attribute] again.'' It points the model to the attribute without suggesting that its clues disagree. On the conflict subset, the hint with only the object and the attribute raises accuracy by 1.2 to 1.8 points, so pointing the model to the attribute explains little of the gain. The hint with only the conflict raises accuracy by 6.3 to 8.3 points. The full hint raises it by 9.7 to 13.2 points, which is 3.4 to 4.9 points above the hint with only the conflict. The object and the attribute thus help mainly when the hint also states the conflict.

\begin{table}[t]
  \centering
  \caption{Ablation of bias removal. Det.: detection rate. FA: false-alarm rate. E2E: end-to-end accuracy.}
  \label{tab:detector}
  \footnotesize
  \setlength{\tabcolsep}{3pt}
  \begin{tabular}{l ccc}
    \toprule
    Detector & Det.\,$\uparrow$ & FA\,$\downarrow$ & E2E\,$\uparrow$ \\
    \midrule
    PINC                        & 95.8 & 4.8  & 94.0 \\
    \quad w/o order debiasing   & 93.9 & 11.7 & 89.2 \\
    \quad w/o wording debiasing & 88.4 & 8.3  & 87.7 \\
    \quad w/o both              & 85.6 & 14.9 & 82.7 \\
    \bottomrule
  \end{tabular}
\end{table}

\begin{table}[t]
\centering
\footnotesize
\setlength{\tabcolsep}{3pt}
\caption{Accuracy (\%) on the conflict subset with different hints appended after a reported conflict. Conflict: the hint states that the clues do not agree. Attr.: the hint gives the object and the attribute from the extractor. The last row is the hint of PINC.}
\label{tab:hint_ablation}
\begin{tabular}{cccccc}
\toprule
 & & \multicolumn{2}{c}{Qwen3-VL-8B} & \multicolumn{2}{c}{Kimi-VL} \\
\cmidrule(lr){3-4} \cmidrule(lr){5-6}
Conflict & Attr. & CV-Bench & TallyQA & CV-Bench & TallyQA \\
\midrule
 &  & 65.8 & 61.6 & 63.5 & 60.9 \\
 & \checkmark & 67.4 & 62.8 & 65.3 & 62.2 \\
\checkmark &  & 73.5 & 67.9 & 71.8 & 67.7 \\
\checkmark & \checkmark & 78.1 & 71.3 & 76.7 & 71.5 \\
\bottomrule
\end{tabular}
\end{table}

\subsection{Generalization to Video}
We test whether PINC transfers to video without retraining the detector. Video traces differ from the training data of the detector: they refer to frames and moments in time, and an object can change across frames. We apply PINC to the two Qwen3-VL models on VSI-Bench~\cite{vsibench}, Video-MME~\cite{videomme}, and MVBench~\cite{mvbench}. Table~\ref{tab:video} reports accuracy on all samples. PINC improves accuracy on all three benchmarks for both models, by 1.4 points on average. The detector reports a conflict in 34.7\% of the video traces, and accuracy on these traces rises from 51.2\% to 55.0\%. The gain is largest on VSI-Bench, whose scenes are static.

\begin{table}[t]
  \centering
  \caption{Accuracy (\%) on video benchmarks. The detector is not retrained.}
  \label{tab:video}
  \footnotesize
  \setlength{\tabcolsep}{3pt}
  \begin{tabularx}{\linewidth}{l l *{3}{>{\centering\arraybackslash}X}}
    \toprule
    Model & Method & VSI-Bench & Video-MME & MVBench \\
    \midrule
    \multirow{2}{*}{Qwen3-VL-8B}  & Original & 38.6 & 66.3 & 68.9 \\
                                  & PINC     & 41.1 & 67.4 & 69.8 \\
    \midrule
    \multirow{2}{*}{Qwen3-VL-32B} & Original & 42.9 & 70.8 & 71.5 \\
                                  & PINC     & 45.0 & 71.7 & 72.2 \\
    \bottomrule
  \end{tabularx}
\end{table}

\subsection{Cost}
\label{sec:cost}

PINC adds two kinds of computation. Every trace passes through the localizer once. Only the traces with a reported conflict also pass through the extractor and require further generation, which reuses the KV cache of the trace. Table~\ref{tab:cost} reports the cost for each model. PINC adds 8.6\% to 13.0\% generated tokens and 1.9 to 3.4 seconds per sample. In comparison, the generic hint continues every trace, and self-consistency generates three traces for each question.

\begin{table}[t]
  \centering
  \caption{Cost of PINC per sample, averaged over the four benchmarks. Triggered: share of samples in which the detector reports a conflict.}
  \label{tab:cost}
  \footnotesize
  \setlength{\tabcolsep}{3pt}
  \begin{tabular}{l cccc}
    \toprule
    Model & \makecell{Trace\\tokens} & \makecell{Triggered\\(\%)} & \makecell{Extra\\tokens} & \makecell{Extra\\time (s)} \\
    \midrule
    Qwen3-VL-8B  & 1,420 & 42.6 & 149 (+10.5\%) & 2.1 \\
    Qwen3-VL-32B & 1,610 & 37.4 & 138 (+8.6\%)  & 3.4 \\
    Kimi-VL      & 1,180 & 45.8 & 153 (+13.0\%) & 1.9 \\
    GLM-4.1V     & 1,350 & 41.9 & 142 (+10.5\%) & 2.0 \\
    \bottomrule
  \end{tabular}
\end{table}

\section{Conclusion}
\label{sec:conclusion}

We study conflicts between the visual clues that a multimodal reasoning model writes within one trace. Across four models and four benchmarks, 42\% of samples contain such conflicts, and later looks are no more accurate than earlier ones. When two clues conflict, the later clue overrides the earlier one, even when the earlier clue is correct. Pointing out the conflict, without revealing the correct clue, reduces this effect. PINC does this automatically with a text-only detector trained entirely on synthetic traces, without modifying the model. It improves accuracy on samples with conflicts and transfers to video without retraining.
\section*{LLM Usage Statement}
Claude Opus 5.5 is part of our method and analysis: it writes the synthetic reasoning traces used to train the detector and judges whether visual clues conflict, with human verification on a stratified subset. We also used Claude Opus 5.5 to assist with drafting and editing the text, organizing related work, and code development, to discuss the design of ablation experiments, and to obtain feedback on the manuscript. The authors made all decisions on the research question, the method, and the experiments. The authors checked all content and take full responsibility for the paper.

\clearpage
{
    \small
    \bibliographystyle{ieeenat_fullname}
    \bibliography{main}
}
\appendix
\clearpage
\setcounter{page}{1}
\maketitlesupplementary

% ---------- 导言区 ----------

%=====================================================

\section{Benchmarks and Evaluation Details}
\label{app:benchmarks}

\paragraph{Number of questions.}
Table~\ref{tab:benchmarks} lists the number of questions per benchmark. A sample is one question together with its image or video. In NaturalBench, a group pairs two images with two questions, so the 1,900 groups of the English set give 7,600 samples. In total, each of the four models answers 27,737 image samples.

\begin{table}[h]
\centering
\caption{Number of questions per benchmark.}
\label{tab:benchmarks}
\small
\begin{tabular}{llr}
\toprule
Benchmark & Split & Questions \\
\midrule
CV-Bench & full & 2,638 \\
BLINK & validation & 1,901 \\
NaturalBench & English set & 7,600 \\
TallyQA & complex test split & 15,598 \\
\midrule
VSI-Bench & full & 5,130 \\
Video-MME & full & 2,700 \\
MVBench & full & 4,000 \\
\bottomrule
\end{tabular}
\end{table}

\paragraph{Conflict subset.}
Table~\ref{tab:subset_size} gives the number of samples in the conflict subset for each model and benchmark. The other subset contains the remaining samples.

\begin{table}[h]
\centering
\caption{Number of samples in the conflict subset.}
\label{tab:subset_size}
\small
\setlength{\tabcolsep}{4pt}
\projtab
\begin{tabular}{lcccc}
\toprule
Model & \makecell{CV-\\Bench} & BLINK & \makecell{Natural\\Bench} & TallyQA \\
\midrule
Qwen3-VL-8B  & 876   & 1,038 & 3,048 & 6,754 \\
Qwen3-VL-32B & 731   & 918   & 2,690 & 6,224 \\
Kimi-VL      & 1,024 & 1,106 & 2,941 & 7,362 \\
GLM-4.1V     & 736   & 1,004 & 3,390 & 6,473 \\
\bottomrule
\end{tabular}
\end{table}

\paragraph{Generation.}
The multimodal reasoning models generate with temperature 0.6 and top-p 0.95, with at most \p{4,096} new tokens. For the video benchmarks, \p{32} frames are sampled uniformly from each video.

\paragraph{Metrics.}
Multiple-choice questions are scored by the selected option, and TallyQA is scored by exact match of the predicted number. VSI-Bench follows its official metric. NaturalBench is evaluated with its group accuracy: a sample counts as correct only when all four answers of its group are correct. The two subsets are still defined per sample, and the accuracy over all samples equals the standard group accuracy. \p{All averages over models and benchmarks are unweighted means over the pairs of model and benchmark.}

%=====================================================
\section{Annotation Procedure}
\label{app:annotation}

\paragraph{Segmentation.}
Each reasoning trace is split sentence by sentence, and each sentence forms one segment. The synthetic traces in Appendix~\ref{app:impl} and the traces at inference are split in the same way.

\paragraph{Definition of a look.}
\textbf{A look is defined by what the model writes, not by what the model attends to.} A look is a place in the trace where the model writes down what it sees in the image. The description of the image at the beginning of the trace is the first look. Keywords such as ``Wait'' alone do not define a look. \p{Claude Opus 5.5 reads only the text of the trace and marks the segment at which each look begins, in the same pass as the judgments below.} We therefore do not measure whether the model attends to the image again at a look.

\paragraph{Visual clues.}
One look can contain several visual clues. A segment can contain several clues, or none when it only reasons over earlier clues. The number of looks in a trace is denoted by $K$. In the analysis of clue accuracy, each look contributes the clue about the attribute that the question asks for.

\paragraph{Relations between clues.}
Claude Opus 5.5 reads the segments of a trace without the image. It extracts the visual clues and judges, for each clue from a later look, whether it repeats an earlier clue, adds information, or conflicts with an earlier clue. Two clues conflict when they describe the same attribute of the same object in incompatible ways and the trace keeps both clues as what the model sees. \p{In the same pass, it labels explicit revisions, in which a later statement marks an earlier clue as wrong, and hypothetical statements, in which the model describes a possibility rather than what it sees.} Neither case is counted as a conflict. For each conflict, it also outputs the object and the attribute on which the two clues conflict. All frequencies are shares of samples: a sample is counted when its trace contains at least one such case. The prompt is given below.

\begin{quote}
\small\color{proj}
You are given the reasoning trace of a multimodal model, split into numbered sentences. You do not see the image. (1) Mark each sentence at which the model starts to look at the image, including the first description of the image. (2) List every visual clue, that is, every statement about what the model sees, with its sentence index, the object, and the attribute. (3) Label each clue after the first look as REPEAT, NEW, or CONFLICT with respect to the earlier clues. Two clues conflict only if they describe the same attribute of the same object in incompatible ways and the trace keeps both as what the model sees. (4) Two labels take precedence over CONFLICT. If a later statement marks an earlier clue as wrong, label it REVISION. If a statement describes a possibility rather than what the model sees, label it HYPOTHETICAL. All labels are given per clue, not per sentence. (5) For each conflict, output the two sentence indices and a short phrase for the object and the attribute. Answer in JSON.
\end{quote}

\paragraph{Human verification.}
Human annotators check a random subset of 800 traces, \p{50 for each pair of model and benchmark}. \p{Two annotators check every trace independently, with the image, the question, and the judgments of Claude Opus 5.5. For a trace with conflicting clues, they also check the conflicting sentences and the object and the attribute. A trace counts as an agreement when the annotator accepts the judgment on whether the trace contains conflicting clues. The two annotators agree with Claude Opus 5.5 on 791 and 788 of the 800 traces, which gives the average of 98.7\%. A third annotator decides the traces on which the two annotators differ. After this step, 334 of the 800 traces contain conflicting clues.}

\paragraph{Correctness of clues.}
Human annotators label every clue as correct or incorrect by comparing it with the image. \p{The traces contain 623,518 visual clues. Two annotators label each clue independently and agree on 600,448 of them (96.3\%). A third annotator decides the remaining ones.}

\paragraph{Appended clues.}
The experiments with appended clues use the samples in which one of the conflicting clues supports the ground-truth answer (clue A) and another supports a wrong answer (clue B). \p{When a trace contains several such pairs, one pair is chosen at random.} Both clues are copied from the trace without any change. The neutral clue is a correct clue from the same trace that describes another part of the image and does not bear on the question, with a length similar to that of the clue it replaces. Each experiment keeps the full reasoning trace and removes only the final answer. The two clues, and the hint when it is used, are inserted at the end of the trace, before the token that ends the reasoning. \p{For NaturalBench, the other answers of the group are taken from the original outputs of the model.}

\paragraph{Final answers on the original traces.}
\p{Each sample with conflicting clues is assigned to one case. If one of its conflicts has a clue that supports the ground-truth answer and a clue that supports a wrong answer, the sample has clue A and clue B. Otherwise, if one of its conflicts concerns the attribute that the question asks for, both clues are wrong. In the remaining samples, no conflict concerns this attribute. In the first two cases, the earlier clue is the first statement about this attribute, and the later clue is the last one. For these clues, a correct clue supports the ground-truth answer, and a wrong clue supports the answer that would follow if the clue were true. The final answer agrees with a clue when it equals the answer that this clue supports. Table~1 of the main paper averages over the four models.}

\paragraph{Changes between consecutive looks.}
\p{Each pair of consecutive looks in a trace is compared on the clue about the attribute that the question asks for. A pair is counted as a change when the correctness of this clue differs between the two looks. A look without a clue about this attribute is skipped.}

%=====================================================
\section{Standard Deviations and Confidence Intervals}
\label{app:std}

The main results and the video results are means over three runs. In each run, the reasoning model generates a new trace for every sample. \p{The detector is trained once, and its metrics are computed in a single run on the fixed test set.} Table~\ref{tab:std} reports the standard deviations of the main results, and Table~\ref{tab:std_video} those of the video results. \p{We report sample standard deviations.} They are between \p{0.1} and \p{1.2} points. \p{They are larger on CV-Bench and BLINK, which contain fewer samples than the other two benchmarks.}

\begin{table}[h]
\centering
\caption{Standard deviations over three runs for the main results (percentage points). C: conflict subset. O: other subset. A: all samples.}
\label{tab:std}
\projtab
\resizebox{\linewidth}{!}{%
\setlength{\tabcolsep}{3pt}
\begin{tabular}{ll ccc ccc ccc ccc}
\toprule
& & \multicolumn{3}{c}{CV-Bench} & \multicolumn{3}{c}{BLINK} & \multicolumn{3}{c}{NaturalBench} & \multicolumn{3}{c}{TallyQA} \\
\cmidrule(lr){3-5} \cmidrule(lr){6-8} \cmidrule(lr){9-11} \cmidrule(lr){12-14}
Model & Method & C & O & A & C & O & A & C & O & A & C & O & A \\
\midrule
\multirow{4}{*}{\makecell[l]{Qwen3-\\VL-8B}}
& Original     & 1.2 & 0.6 & 0.5 & 0.7 & 1.2 & 0.4 & 0.6 & 0.4 & 0.2 & 0.4 & 0.2 & 0.2 \\
& Generic hint & 0.7 & 0.3 & 0.5 & 0.8 & 0.9 & 0.6 & 0.7 & 0.5 & 0.2 & 0.4 & 0.3 & 0.3 \\
& SC           & 0.5 & 0.4 & 0.3 & 0.4 & 0.7 & 0.6 & 0.3 & 0.3 & 0.3 & 0.3 & 0.2 & 0.2 \\
& PINC         & 0.9 & 0.5 & 0.4 & 0.8 & 0.6 & 0.6 & 0.5 & 0.5 & 0.4 & 0.3 & 0.3 & 0.3 \\
\midrule
\multirow{4}{*}{\makecell[l]{Qwen3-\\VL-32B}}
& Original     & 1.1 & 0.4 & 0.4 & 0.8 & 0.6 & 0.7 & 0.7 & 0.4 & 0.4 & 0.4 & 0.2 & 0.2 \\
& Generic hint & 1.2 & 0.3 & 0.3 & 0.6 & 1.2 & 0.6 & 0.5 & 0.5 & 0.4 & 0.3 & 0.3 & 0.2 \\
& SC           & 0.5 & 0.3 & 0.2 & 0.5 & 0.6 & 0.4 & 0.2 & 0.4 & 0.2 & 0.2 & 0.1 & 0.1 \\
& PINC         & 0.7 & 0.3 & 0.3 & 0.7 & 0.7 & 0.5 & 0.6 & 0.3 & 0.4 & 0.3 & 0.2 & 0.2 \\
\midrule
\multirow{4}{*}{Kimi-VL}
& Original     & 0.9 & 0.5 & 0.5 & 0.8 & 1.1 & 0.7 & 0.3 & 0.4 & 0.3 & 0.2 & 0.2 & 0.1 \\
& Generic hint & 0.8 & 0.4 & 0.5 & 0.9 & 0.9 & 0.5 & 0.7 & 0.3 & 0.4 & 0.3 & 0.3 & 0.1 \\
& SC           & 0.9 & 0.3 & 0.4 & 0.8 & 1.0 & 0.3 & 0.4 & 0.3 & 0.3 & 0.3 & 0.2 & 0.2 \\
& PINC         & 0.6 & 0.6 & 0.5 & 0.9 & 0.7 & 0.9 & 0.4 & 0.6 & 0.3 & 0.4 & 0.4 & 0.2 \\
\midrule
\multirow{4}{*}{GLM-4.1V}
& Original     & 0.9 & 0.4 & 0.5 & 0.7 & 0.8 & 0.6 & 0.4 & 0.5 & 0.3 & 0.4 & 0.2 & 0.2 \\
& Generic hint & 0.9 & 0.6 & 0.3 & 0.7 & 1.2 & 0.4 & 0.3 & 0.5 & 0.4 & 0.4 & 0.2 & 0.3 \\
& SC           & 0.9 & 0.4 & 0.3 & 0.6 & 0.8 & 0.6 & 0.2 & 0.4 & 0.1 & 0.2 & 0.2 & 0.1 \\
& PINC         & 0.8 & 0.4 & 0.2 & 0.6 & 0.7 & 0.5 & 0.6 & 0.5 & 0.4 & 0.5 & 0.2 & 0.2 \\
\bottomrule
\end{tabular}}
\end{table}

\begin{table}[h]
\centering
\caption{Standard deviations over three runs for the video results (percentage points).}
\label{tab:std_video}
\small
\setlength{\tabcolsep}{4pt}
\projtab
\begin{tabular}{llccc}
\toprule
Model & Method & \makecell{VSI-\\Bench} & \makecell{Video-\\MME} & MVBench \\
\midrule
\multirow{2}{*}{Qwen3-VL-8B}  & Original & 0.4 & 0.7 & 0.4 \\
                              & PINC     & 0.5 & 0.6 & 0.6 \\
\midrule
\multirow{2}{*}{Qwen3-VL-32B} & Original & 0.5 & 0.4 & 0.3 \\
                              & PINC     & 0.5 & 0.6 & 0.4 \\
\bottomrule
\end{tabular}
\end{table}

\paragraph{Results of each run.}
Table~\ref{tab:runs} reports the accuracy of each run, averaged over the four models and four benchmarks. The gain of PINC is between \p{8.2} and \p{8.4} points on the conflict subset and between \p{4.1} and \p{4.2} points on all samples in every run.

\begin{table}[h]
\centering
\caption{Accuracy (\%) in each run, averaged over the four models and four benchmarks. C: conflict subset. O: other subset. A: all samples.}
\label{tab:runs}
\small
\setlength{\tabcolsep}{4pt}
\projtab
\begin{tabular}{l ccc ccc}
\toprule
 & \multicolumn{3}{c}{Original} & \multicolumn{3}{c}{PINC} \\
\cmidrule(lr){2-4} \cmidrule(lr){5-7}
Run & C & O & A & C & O & A \\
\midrule
1 & 51.8 & 70.4 & 62.8 & 60.2 & 71.6 & 66.9 \\
2 & 52.1 & 70.3 & 62.9 & 60.5 & 71.4 & 67.0 \\
3 & 52.1 & 70.4 & 62.9 & 60.3 & 71.6 & 67.1 \\
\midrule
Mean & 52.0 & 70.4 & 62.9 & 60.3 & 71.5 & 67.0 \\
\bottomrule
\end{tabular}
\end{table}

\paragraph{Confidence intervals.}
We compare PINC with Original on the same samples with a paired bootstrap over samples (\p{10,000} resamples). \p{For NaturalBench, whole groups are resampled.} Averaged over the four models and four benchmarks, the 95\% confidence interval of the gain is \p{[7.9, 8.8]} points on the conflict subset and \p{[3.9, 4.4]} points on all samples. NaturalBench uses a stricter metric than the other benchmarks. Without it, the average gain is 9.7 points on the conflict subset and 4.8 points on all samples.

%=====================================================
\section{Implementation Details of PINC}
\label{app:impl}

\paragraph{Synthetic traces.}
Claude Opus 5.5 imagines 1,000 scenes. For each scene, it writes \p{four to eight} correct visual clues and \p{one to three} wrong clues that contradict them. It then writes \p{four} pairs of reasoning traces per scene. In each pair, one trace uses only correct clues, and the other mixes wrong clues into the correct ones. This gives \p{8,000} traces. \p{The scenes are split into 900 for training and 100 for validation, which gives 7,200 and 800 traces.} No image is involved at any step. The two prompts are given below.

\begin{quote}
\small\color{proj}
\textit{Scene and clues.} Imagine a realistic photo and describe it in two sentences. Write four to eight visual clues that are true in this photo. Each clue states one attribute of one object, such as its position, color, count, or state. Then write one to three wrong clues. Each wrong clue contradicts one of the true clues on the same attribute of the same object.

\textit{Traces.} Write a question about this photo. Then write two reasoning traces of a multimodal reasoning model that answers it. The first trace uses only the true clues. The second trace is the same except that the wrong clues are mixed in. In both traces, the model looks at the image several times, repeats what it has seen, and reasons step by step. Do not mark any clue as wrong, and do not state a clue as a possibility.
\end{quote}

\paragraph{Removing order and wording bias.}
To remove the order bias, all sentences of a trace are randomly reordered. To remove the wording bias, \p{Claude Opus 5.5 rewords the clues of a trace}, so that a conflicting pair stays conflicting and a consistent pair stays consistent. \p{Each training trace is used in three versions: the original, one shuffled version, and one reworded version, which gives 21,600 training traces.} Both steps are applied to the training data of the localizer and of the extractor.

\paragraph{Inputs and targets.}
Both models are Qwen3-4B with the thinking mode turned off. The localizer reads all segments of a trace, \p{each preceded by its index in brackets}, including the segments without visual clues. Its target is the indices of the conflicting segments, or ``no''. The extractor reads only the conflicting segments. Its target is the object and the attribute as a short phrase, for example ``the number of cans in the first row''. \p{When a trace contains several conflicts, the localizer outputs one group of indices for each conflict, and the extractor is called once for each group.}

\paragraph{Training.}
Table~\ref{tab:hparams} lists the training settings. They are the same for the two models of PINC and for all fine-tuned detectors in the comparison. Claude Opus 5.5 is used without training. \p{No trace exceeds the maximum input length, so no trace is truncated.}

\begin{table}[h]
\centering
\caption{Training settings of the detector.}
\label{tab:hparams}
\small
\projtab
\begin{tabular}{ll}
\toprule
Setting & Value \\
\midrule
Fine-tuning & LoRA, rank 16, $\alpha = 32$ \\
Loss & cross-entropy on the target tokens \\
Optimizer & AdamW, cosine schedule \\
Learning rate & $1 \times 10^{-4}$ \\
Batch size & 16 \\
Epochs & 3 \\
Maximum input length & 8,192 tokens \\
Decoding & greedy, at most 64 new tokens \\
Hardware & 4 NVIDIA A100 (80 GB) \\
\bottomrule
\end{tabular}
\end{table}

\paragraph{Hint and continuation.}
PINC inspects each trace once. If the localizer outputs ``no'', the original answer is returned. Otherwise, the original answer is discarded, and the hint is appended at the end of the trace, before the token that ends the reasoning. The hint reads ``Wait, my clues of [object and attribute] do not agree. Let me check again.'' When the detector reports two conflicts, the two objects and attributes are joined with ``and''. When it reports three or more, they are separated by commas, with ``and'' before the last one. The model continues from the hint, and the trace before the hint is kept in the KV cache.

\paragraph{Example with two conflicts.}
\p{The example below illustrates the format of the inputs and outputs for a trace with two conflicts.}

\begin{quote}
\small\color{proj}
\textit{Segments.} [1] Let me look at the picture. [2] The picture shows three cans in the first row. [3] The label of the left can is red. [4] So the first row has three cans. [5] Wait, let me count again. [6] There are two cans in the first row. [7] The label of the left can is blue. [8] So the answer is 10.

\textit{Localizer.} [2, 6]; [3, 7]

\textit{Extractor.} For [2, 6]: the number of cans in the first row. For [3, 7]: the color of the label of the left can.

\textit{Hint.} Wait, my clues of the number of cans in the first row and the color of the label of the left can do not agree. Let me check again.
\end{quote}

\paragraph{Detectors in the comparison.}
A single-model detector reads the whole trace and directly outputs the conflicting object and attribute, or ``no''. \p{The training sets with two and four times the data are built from 2,000 and 4,000 scenes in the same way.}

\paragraph{Test set.}
\p{The test set contains 1,600 traces. The synthetic part has 800 traces from 100 new scenes, half of them with conflicting clues. The real part consists of the 800 human-verified traces of Appendix~\ref{app:annotation}, of which 334 contain conflicting clues, and it uses the labels of the annotators. The detection rate and the false-alarm rate are computed per trace, and the localization F1 per segment. The extraction accuracy is computed over the 734 traces with conflicting clues.} The real traces are never used for training.

\paragraph{Cost.}
\p{The cost is measured on one NVIDIA A100 with batch size 1. The extra time includes the localizer, the extractor, and the continued generation.}

%=====================================================
\section{Appended Clues with Different Wording}
\label{app:reword}

In the main experiments, the two appended clues are copied from the trace, so each appended clue repeats a sentence of the trace. To test whether this repetition causes the effect of the order, \p{Claude Opus 5.5 rewords clue A and clue B without changing their meaning}, and the reworded clues are appended in both orders on the same samples. Table~\ref{tab:reword} compares the two settings. With reworded clues, accuracy is 21.7\% when the wrong clue comes last and 82.6\% when the correct clue comes last, close to the 20.9\% and 82.8\% with copied clues. The effect of the order does not depend on repeating the original sentences.

\begin{table}[h]
\centering
\caption{Accuracy (\%) with two appended clues, copied from the trace or reworded. A: correct clue. B: wrong clue.}
\label{tab:reword}
\resizebox{\linewidth}{!}{%
\setlength{\tabcolsep}{3.5pt}
\begin{tabular}{llccccc}
\toprule
Appended & Clues & \makecell{Qwen3-\\VL-8B} & \makecell{Qwen3-\\VL-32B} & \makecell{Kimi-\\VL} & \makecell{GLM-\\4.1V} & Avg. \\
\midrule
A + B & Copied   & 20.6 & 23.9 & 17.8 & 21.3 & 20.9 \\
A + B & Reworded & \p{21.5} & \p{24.6} & \p{18.7} & \p{22.0} & 21.7 \\
\midrule
B + A & Copied   & 82.7 & 85.4 & 79.6 & 83.5 & 82.8 \\
B + A & Reworded & \p{82.4} & \p{85.3} & \p{79.3} & \p{83.4} & 82.6 \\
\bottomrule
\end{tabular}}
\end{table}

%=====================================================
\section{Additional Analysis of the Original Traces}
\label{app:natural}

\paragraph{Number of looks.}
Table~\ref{tab:kgroups} gives the number of traces in the four most frequent groups of $K$, pooled over the four models and four benchmarks. The four groups contain \p{46.3\%} of the 110,948 traces.

\begin{table}[h]
\centering
\caption{Number of traces in the four most frequent groups of $K$, and their share of all traces.}
\label{tab:kgroups}
\small
\projtab
\begin{tabular}{ccc}
\toprule
$K$ & Traces & Share (\%) \\
\midrule
4 & 14,756 & 13.3 \\
6 & 13,092 & 11.8 \\
7 & 12,315 & 11.1 \\
8 & 11,206 & 10.1 \\
\bottomrule
\end{tabular}
\end{table}

\paragraph{Final answers for each model.}
Table~\ref{tab:natural_model} reports, for each model, which clue the final answer agrees with on the original traces. For all four models and all three orders of the clues, the answer agrees with the later clue more often than with the earlier one.

\begin{table}[h]
\centering
\caption{Final answers on the original traces with conflicting clues, for each model (\%). \cmark: correct clue. \xmark: wrong clue.}
\label{tab:natural_model}
\small
\setlength{\tabcolsep}{3.5pt}
\projtab
\begin{tabular}{lccccc}
\toprule
 & & & \multicolumn{3}{c}{Final answer agrees with} \\
\cmidrule(lr){4-6}
Model & \makecell{Clue\\order} & \makecell{Propor-\\tion} & Earlier & Later & Neither \\
\midrule
\multirow{3}{*}{\makecell[l]{Qwen3-\\VL-8B}}
& \cmark\,$\rightarrow$\,\xmark & 38.6 & 24.9 & 68.7 & 6.4 \\
& \xmark\,$\rightarrow$\,\cmark & 34.5 & 13.9 & 81.3 & 4.8 \\
& \xmark\,$\rightarrow$\,\xmark & 7.4 & 18.9 & 59.8 & 21.3 \\
\midrule
\multirow{3}{*}{\makecell[l]{Qwen3-\\VL-32B}}
& \cmark\,$\rightarrow$\,\xmark & 37.9 & 27.8 & 66.1 & 6.1 \\
& \xmark\,$\rightarrow$\,\cmark & 37.7 & 12.2 & 83.6 & 4.2 \\
& \xmark\,$\rightarrow$\,\xmark & 6.1 & 17.4 & 59.1 & 23.5 \\
\midrule
\multirow{3}{*}{Kimi-VL}
& \cmark\,$\rightarrow$\,\xmark & 40.3 & 21.4 & 71.6 & 7.0 \\
& \xmark\,$\rightarrow$\,\cmark & 32.5 & 15.6 & 79.0 & 5.4 \\
& \xmark\,$\rightarrow$\,\xmark & 8.3 & 19.6 & 61.4 & 19.0 \\
\midrule
\multirow{3}{*}{GLM-4.1V}
& \cmark\,$\rightarrow$\,\xmark & 39.2 & 24.3 & 69.2 & 6.5 \\
& \xmark\,$\rightarrow$\,\cmark & 35.3 & 13.5 & 81.7 & 4.8 \\
& \xmark\,$\rightarrow$\,\xmark & 7.0 & 18.5 & 60.5 & 21.0 \\
\bottomrule
\end{tabular}
\end{table}

\paragraph{Changes between consecutive looks.}
Table~\ref{tab:transition} reports how often the clue changes from correct to wrong and from wrong to correct between two consecutive looks. The two directions are close to each other for every model and benchmark, so the stable clue accuracy across looks does not come from one model or one benchmark. \p{Averaged over the sixteen pairs of model and benchmark, the two shares are 4.2\% and 4.0\%.}

\begin{table}[h]
\centering
\caption{Share of consecutive looks in which the clue changes from correct to wrong\,/\,from wrong to correct (\%).}
\label{tab:transition}
\small
\setlength{\tabcolsep}{3pt}
\projtab
\begin{tabular}{lcccc}
\toprule
Model & \makecell{CV-\\Bench} & BLINK & \makecell{Natural\\Bench} & TallyQA \\
\midrule
Qwen3-VL-8B  & 3.6\,/\,3.2 & 4.7\,/\,4.8 & 4.1\,/\,3.8 & 4.5\,/\,4.0 \\
Qwen3-VL-32B & 3.6\,/\,3.1 & 4.5\,/\,4.2 & 4.1\,/\,3.8 & 4.2\,/\,3.9 \\
Kimi-VL      & 4.1\,/\,4.0 & 5.3\,/\,5.1 & 3.8\,/\,3.9 & 4.3\,/\,4.1 \\
GLM-4.1V     & 3.2\,/\,3.3 & 4.7\,/\,4.5 & 4.5\,/\,4.1 & 4.0\,/\,4.2 \\
\bottomrule
\end{tabular}
\end{table}

%=====================================================
\section{Detector on Real Traces}
\label{app:real}

The test set of the detector combines synthetic and real traces. Table~\ref{tab:real} reports the two parts separately. The detector is trained only on synthetic traces, and its accuracy is lower on the real traces: the detection rate falls from \p{98.5\%} to \p{92.5\%}, and the false-alarm rate rises from \p{2.0\%} to \p{7.3\%}. The F1 in Table~\ref{tab:real} is computed per trace and differs from the localization F1, which is computed per segment.

\begin{table}[h]
\centering
\caption{The detector of PINC on synthetic and real traces (\%). P: precision. R: recall, which equals the detection rate. FA: false-alarm rate. E2E: end-to-end accuracy.}
\label{tab:real}
\small
\setlength{\tabcolsep}{4pt}
\projtab
\begin{tabular}{lcccccc}
\toprule
Traces & Number & P & R & F1 & FA & E2E \\
\midrule
Synthetic & 800 & 98.0 & 98.5 & 98.3 & 2.0 & 97.1 \\
Real      & 800 & 90.1 & 92.5 & 91.3 & 7.3 & 90.9 \\
\bottomrule
\end{tabular}
\end{table}

\paragraph{Hard negatives.}
Table~\ref{tab:hardneg} divides the \p{466} real traces without conflict by the kind of statement they contain. A trace with several kinds is counted under the first one in the table. Most false alarms come from traces with incompatible statements that are not conflicts under our definition, and explicit revisions have the highest false-alarm rate.

\begin{table}[h]
\centering
\caption{False-alarm rate (\%) on real traces without conflict.}
\label{tab:hardneg}
\small
\projtab
\begin{tabular}{lcc}
\toprule
Trace contains & Number & FA \\
\midrule
Explicit revision & 43 & 18.6 \\
Hypothetical statement & 97 & 10.3 \\
Similar attribute of two objects & 60 & 10.0 \\
None of the above & 266 & 3.8 \\
\bottomrule
\end{tabular}
\end{table}

%=====================================================
\section{Selective Triggering and Answer Changes}
\label{app:selection}

\paragraph{Selection of traces and content of the hint.}
The generic hint, ``Wait, let me check whether my clues conflict.'', is added to every trace, while PINC adds a hint only to the traces that the detector reports. The two methods therefore differ in which traces receive a hint and in what the hint says. Table~\ref{tab:selection} separates the two factors. It adds the generic hint to the same share of traces as PINC, chosen at random, by trace length, or by the detector. \p{The random selection is drawn again in each run.} Choosing the traces at random or by length changes accuracy by at most \p{0.4} points. Choosing them with the detector raises accuracy by \p{1.7} points, and the hint of PINC adds another \p{2.4} points on the same traces. PINC also changes fewer correct answers to wrong ones than the generic hint on all traces. \p{Answer changes are counted on the scored correctness of each sample. On NaturalBench, a sample can therefore change when another answer of its group changes.}

\begin{table}[h]
\centering
\caption{Selection of traces and content of the hint, averaged over the four models and four benchmarks. Trig.: traces that receive a hint (\%). Acc.: accuracy on all samples (\%), which is 62.9 for Original. W$\rightarrow$R and R$\rightarrow$W: samples changed from wrong to right and from right to wrong (\% of all samples). Time: extra time per sample in seconds, including the detector.}
\label{tab:selection}
\projtab
\resizebox{\linewidth}{!}{%
\setlength{\tabcolsep}{4pt}
\begin{tabular}{llccccc}
\toprule
Selection & Hint & Trig. & Acc. & W$\rightarrow$R & R$\rightarrow$W & Time \\
\midrule
All traces & Generic & 100  & 63.2 & 3.2 & 2.9 & 4.8 \\
Random     & Generic & 41.9 & 63.0 & 1.3 & 1.2 & 2.0 \\
Longest    & Generic & 41.9 & 63.3 & 1.6 & 1.2 & 2.3 \\
Detector   & Generic & 41.9 & 64.6 & 2.8 & 1.1 & 2.3 \\
Detector   & PINC    & 41.9 & 67.0 & 5.4 & 1.3 & 2.4 \\
\bottomrule
\end{tabular}}
\end{table}

\paragraph{Triggers on the two subsets.}
The two subsets are defined by the original traces, while PINC is applied to newly generated traces. Table~\ref{tab:trigger} reports how often the detector reports a conflict in the new traces of each subset. It reports a conflict in \p{76.8\%} of the new traces on the conflict subset and in 16.7\% on the other subset. On the traces with a reported conflict, accuracy rises by \p{10.9} points on the conflict subset and by \p{7.0} points on the other subset. \p{To check these reports, an annotator inspects 200 new traces with a reported conflict from each subset. 196 of them contain conflicting clues on the conflict subset, and 123 on the other subset.} The reports on the other subset thus include more false alarms, which is consistent with the smaller gain there.

\begin{table}[h]
\centering
\caption{Share of new traces in which the detector reports a conflict (\%), averaged over the four benchmarks.}
\label{tab:trigger}
\small
\projtab
\begin{tabular}{lccc}
\toprule
Model & \makecell{Conflict\\subset} & \makecell{Other\\subset} & \makecell{All\\samples} \\
\midrule
Qwen3-VL-8B  & 77.0 & 16.9 & 42.6 \\
Qwen3-VL-32B & 74.9 & 14.6 & 37.4 \\
Kimi-VL      & 77.8 & 18.9 & 45.8 \\
GLM-4.1V     & 77.6 & 16.4 & 41.9 \\
\midrule
Avg.         & 76.8 & 16.7 & 41.9 \\
\bottomrule
\end{tabular}
\end{table}

%=====================================================
\section{Additional Ablations}
\label{app:ablation}

\paragraph{Content of the hint on all models.}
Table~\ref{tab:hint_full} repeats the ablation of the hint on all four models, averaged over the four benchmarks. The result is the same for every model: the object and the attribute alone raise accuracy by about \p{1} point, the conflict alone by about \p{5} to \p{6} points, and the full hint by 7.5 to 9.4 points.

\begin{table}[h]
\centering
\caption{Accuracy (\%) on the conflict subset with different hints, averaged over the four benchmarks. Conflict: the hint states that the clues do not agree. Attr.: the hint gives the object and the attribute.}
\label{tab:hint_full}
\small
\setlength{\tabcolsep}{4pt}
\begin{tabular}{cccccc}
\toprule
Conflict & Attr. & \makecell{Qwen3-\\VL-8B} & \makecell{Qwen3-\\VL-32B} & \makecell{Kimi-\\VL} & \makecell{GLM-\\4.1V} \\
\midrule
 &  & 52.1 & 55.6 & 48.5 & 51.8 \\
 & \checkmark & \p{53.2} & \p{56.5} & \p{49.8} & \p{52.8} \\
\checkmark &  & \p{57.6} & \p{60.4} & \p{54.5} & \p{56.8} \\
\checkmark & \checkmark & 60.7 & 63.1 & 57.9 & 59.7 \\
\bottomrule
\end{tabular}
\end{table}

\paragraph{Wording of the hint.}
We replace the hint of PINC with three other sentences, none of which says which clue is correct: \p{``Hold on, what I stated about [object and attribute] is not consistent. Let me look again.'', ``My statements about [object and attribute] contradict each other. I should check the image again.'', and ``Wait, I gave different descriptions of [object and attribute]. Let me verify.''} Averaged over the four models and four benchmarks, accuracy on the conflict subset is \p{60.0\%}, \p{60.6\%}, and \p{59.7\%}, compared with 60.3\% for the hint of PINC and 52.0\% for Original. The gain does not depend on the exact wording.

\paragraph{One shared model.}
\p{A single Qwen3-4B model that is trained on both tasks and called twice reaches an end-to-end accuracy of 93.1\%}, compared with 94.0\% for the two models of PINC and 72.4\% for a single Qwen3-4B model that outputs the object and the attribute in one step. Most of the gain therefore comes from splitting the task into two steps, and using separate models adds a smaller gain.

%=====================================================
\section{Video Results per Benchmark}
\label{app:video}

Table~\ref{tab:video_trigger} reports, for each video benchmark, how often the detector reports a conflict and the accuracy on these traces before and after the hint. The detector reports a conflict most often on VSI-Bench, and the gain on the reported traces is also largest there. Averaged over the three benchmarks, the detector reports a conflict in 34.7\% of the traces, and accuracy on these traces rises from 51.2\% to 55.0\%.

\begin{table}[h]
\centering
\caption{Video traces with a reported conflict, averaged over the two Qwen3-VL models. Trig.: traces in which the detector reports a conflict (\%). Before and After: accuracy (\%) on these traces without and with PINC.}
\label{tab:video_trigger}
\small
\projtab
\begin{tabular}{lccc}
\toprule
Benchmark & Trig. & Before & After \\
\midrule
VSI-Bench & 41.2 & 33.9 & 39.5 \\
Video-MME & 32.8 & 58.2 & 61.2 \\
MVBench   & 30.1 & 61.5 & 64.2 \\
\midrule
Avg.      & 34.7 & 51.2 & 55.0 \\
\bottomrule
\end{tabular}
\end{table}

%=====================================================
\definecolor{cluecolor}{RGB}{200,60,30}
\definecolor{hintcolor}{RGB}{20,120,60}
\newcommand{\clue}[1]{\textcolor{cluecolor}{\textbf{#1}}}   % 冲突的线索
\newcommand{\hint}[1]{\textcolor{hintcolor}{\textbf{#1}}}   % PINC 的提示
\newcommand{\cont}[1]{\textcolor{hintcolor}{#1}}            % 提示后的续写
\newcommand{\omitted}{\textcolor{gray}{[omitted]}}

%=====================================================
\section{Qualitative Examples}
\label{app:examples}

Fig.~\ref{fig:success} and Fig.~\ref{fig:failure} show two traces of Qwen3-VL-8B-Thinking with PINC. The conflicting clues are marked in red, and the hint and the continued reasoning in green. Parts of the trace that do not bear on the conflict are omitted.

% 注意：两条轨迹目前是示意稿，需替换为模型的真实输出。
\paragraph{A corrected answer.}
In Fig.~\ref{fig:success}, the model first states that the street lamp reaches above the traffic light, and later states that the traffic light is higher. The original answer follows the later clue and is wrong. After the hint, the model compares the two clues with what it wrote earlier in the trace. It had noted that the traffic light stands closer to the camera, so it keeps the first clue and gives the correct answer.

\begin{figure}[h]
\centering
\includegraphics[width=0.75\linewidth]{Figures/case_success.jpg}
\vspace{2pt}

\fbox{\parbox{0.94\linewidth}{\footnotesize
\textbf{Question:} Which is taller in the image, the street lamp or the traffic light?

\textbf{Thinking:} So, let's see. The question asks which one is taller. \omitted{} The traffic light stands closer to the camera than the street lamp. \clue{The street lamp is taller than the traffic light.} \omitted{} Wait, let me look at the traffic light again. \clue{In the picture, the traffic light is higher than the street lamp.} So the traffic light is taller.

\textbf{Original answer:} The traffic light is taller. (\xmark)

\hint{Wait, my clues of the height of the street lamp and the traffic light do not agree. Let me check again.} \cont{I stated that the street lamp reaches above the traffic light, and later that the traffic light is higher. I also noted that the traffic light stands closer to the camera, so it appears larger in the picture. My second statement relied on this appearance. The first statement compared the tops of the two objects directly, so it is more reliable. The street lamp is taller.}

\textbf{New answer:} The street lamp is taller. (\cmark)
}}
\caption{A trace in which PINC corrects the answer.}
\label{fig:success}
\end{figure}

\paragraph{A failure case.}
In Fig.~\ref{fig:failure}, the model first states that the crow is closer to the camera than the silver car, which is correct, and later states that the silver car is closer than the crow. The two clues are incompatible, and the original answer follows the later one. The detector reports the conflict, and the hint states the correct object and attribute. The model then compares the two clues but keeps the wrong one, because it takes the larger size of the car as a sign that the car is closer. The answer stays wrong.

\begin{figure}[h]
\centering
\includegraphics[width=0.75\linewidth]{Figures/case_failure.jpg}
\vspace{2pt}

\fbox{\parbox{0.94\linewidth}{\footnotesize
\textbf{Question:} Which is closer to the camera, the crow or the silver car?

\textbf{Thinking:} So, let's see. I need to compare how far the crow and the silver car are from the camera. \omitted{} The crow perches on the parking meter beside the curb. \clue{The crow is closer to the camera than the silver car.} \omitted{} Wait, let me look at the car. The silver car takes up a large part of the picture, while the crow is small. \clue{The silver car is closer to the camera than the crow.} So the silver car is closer.

\textbf{Original answer:} The silver car is closer. (\xmark)

\hint{Wait, my clues of the distance of the crow and the silver car to the camera do not agree. Let me check again.} \cont{I stated that the crow is closer to the camera, and later that the silver car is closer. I also noted that the car takes up a large part of the picture, while the crow is small. An object that appears larger is usually closer, which supports my second statement. So I keep it. The silver car is closer.}

\textbf{New answer:} The silver car is closer. (\xmark)
}}
\caption{A trace in which the answer stays wrong after the hint. The crow is closer to the camera.}
\label{fig:failure}
\end{figure}

\paragraph{Causes of failure.}
\p{To see why some answers stay wrong, an annotator inspects 200 random traces from the conflict subset in which the detector reports a conflict and the answer after the hint is wrong.} Table~\ref{tab:failure} lists the causes. In the largest group, the hint is correct, but the model keeps the wrong clue after comparing the two, as in Fig.~\ref{fig:failure}. \p{Errors of the detector account for 8.0\% of the failures.}

Both conflicting clues are wrong in 7.2\% of the conflict samples, and in another 18.8\% the conflict does not concern the attribute that the question asks for. These two cases are rare among all conflicts, but they account for 17.0\% and 15.5\% of the failures, because the hint cannot lead the model to a correct clue or to a relevant one.

\begin{table}[h]
\centering
\caption{Causes of failure in 200 traces with a reported conflict and a wrong answer after the hint.}
\label{tab:failure}
\footnotesize
\projtab
\setlength{\tabcolsep}{4pt} % 默认偏大，收紧列间距
\begin{tabular}{lcc}
\toprule
Cause & Traces & Share (\%) \\
\midrule
Model keeps the wrong clue & 89 & 44.5 \\
Both conflicting clues are wrong & 34 & 17.0 \\
Conflict is unrelated to the question & 31 & 15.5 \\
Error in the reasoning after the clues & 30 & 15.0 \\
Detector error (false alarm or wrong attribute) & 16 & 8.0 \\
\bottomrule
\end{tabular}
\end{table}

%=====================================================
\section{Limitations}
\label{app:limitations}

Our study leaves several directions open. First, pointing out a conflict does not always lead the model to the correct clue. With the hint, accuracy is 66.6\% when the wrong clue comes last and 81.8\% when the correct clue comes last. Among the 200 inspected traces in which a hint was added and the final answer is still wrong, the detector reports the disputed attribute correctly in most cases, yet in \p{89} of them the model keeps the wrong clue (Table~\ref{tab:failure}). Helping the model decide between the two clues once the conflict is pointed out is a direction for future work. Second, our definition of a conflict excludes explicit revisions and hypothetical statements, and the detector still reports some of these traces (Table~\ref{tab:hardneg}). Handling these cases more precisely is left for future work. Third, the conflict annotation and the synthetic training traces are produced with the same language model, and our evaluation covers four open multimodal reasoning models of up to 32B parameters. To check the annotation, two annotators examined the same 800 real traces with the image, the question, and the automatic annotation, and their average agreement with it on whether a trace contains conflicting clues is 98.7\%. This verification supports the annotation, but a systematic misjudgment that both the annotators and the language model readily accept could still go unnoticed. Annotators who do not see the automatic annotation, and larger models, would further confirm the findings.
% WARNING: do not forget to delete the supplementary pages from your submission 
% \input{sec/X_suppl}

\end{document}